\documentclass[10pt,twocolumn,letterpaper]{article}

\usepackage[pagenumbers]{cvpr} 

\usepackage{graphicx}
\usepackage{amsmath}
\usepackage{amssymb}
\usepackage{booktabs}

\usepackage[pagebackref,breaklinks,colorlinks]{hyperref}
\usepackage[ruled,vlined]{algorithm2e}

\usepackage[capitalize]{cleveref}
\crefname{section}{Sec.}{Secs.}
\Crefname{section}{Section}{Sections}
\Crefname{table}{Table}{Tables}
\crefname{table}{Tab.}{Tabs.}

\def\cvprPaperID{*****} 
\def\confName{CVPR}
\def\confYear{2022}

\begin{document}

\title{Fast Online and Relational Tracking}

\author{Mohammad Hossein Nasseri$^1$\\
{\tt\small mhnasseri@ut.ac.ir}
\and
Mohammadreza Babaee$^2$\\
{\tt\small reza.babaee@tum.de}
\and
Hadi Moradi$^1$\\
{\tt\small moradih@ut.ac.ir}
\and
Reshad Hosseini$^1$\\
{\tt\small reshad.hosseini@ut.ac.ir}
\and 
$^1$School of Electrical and Computer Engineering, University of Tehran, Tehran, Iran
\and 
$^2$Independent researcher\\
}
\maketitle

\begin{abstract}
To overcome challenges in multiple object tracking task, recent algorithms use interaction cues alongside motion and appearance features. These algorithms use graph neural networks or transformers to extract interaction features that lead to high computation costs. In this paper, a novel interaction cue based on geometric features is presented aiming to detect occlusion and re-identify lost targets with low computational cost. 
Moreover, in most algorithms, camera motion is considered negligible, which is a strong assumption that is not always true and leads to ID Switch or mismatching of targets. In this paper, a method for measuring camera motion and removing its effect is presented that efficiently reduces the camera motion effect on tracking. 
The proposed algorithm is evaluated on MOT17 and MOT20 datasets and it achieves the state-of-the-art performance of MOT17 and comparable results on MOT20. The code is also publicly available\footnote{https://github.com/mhnasseri/for\_tracking}.
\end{abstract}

\section{Introduction}
\label{sec:intro}

Multiple Object Tracking (MOT) is a widely researched and challenging task in computer vision and video surveillance. It is a hard problem to solve due to the similarity of targets, existence of occlusion of targets, entering new targets, and exiting targets~\cite{milan2016mot16}. The most used paradigm to solve this problem is tracking-by-detection~\cite{babaee2017combined}. In this paradigm, first, a detection algorithm localizes targets in a frame and generates a set of bounding boxes. Next, a tracking algorithm assigns an identity to them by associating detections in the current frame with predicted targets in the current frame. The tracking algorithm computes a similarity score between detections and tracklets and a matching algorithm such as Hungarian algorithm ~\cite{bewley2016simple,fang2018recurrent}, network flow~\cite{babaee2016pixel,babaee2017combined,dehghan2015target,braso2020learning}, and multiple hypothesis tracking~\cite{chen2017enhancing,kim2015multiple} use this score for the assignment. 

\begin{figure*}[t]
\centering
\begin{tabular}{c c}
\includegraphics[width=7.5cm, height=4cm]{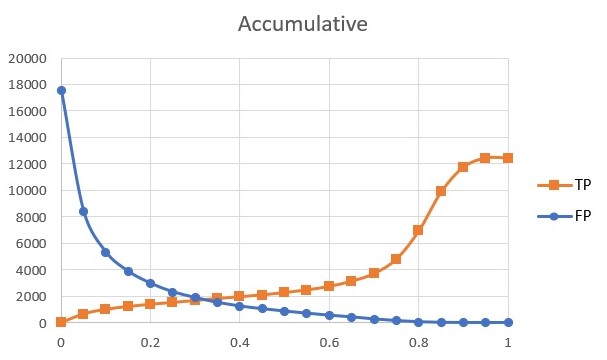}&
\includegraphics[width=7.5cm, height=4cm]{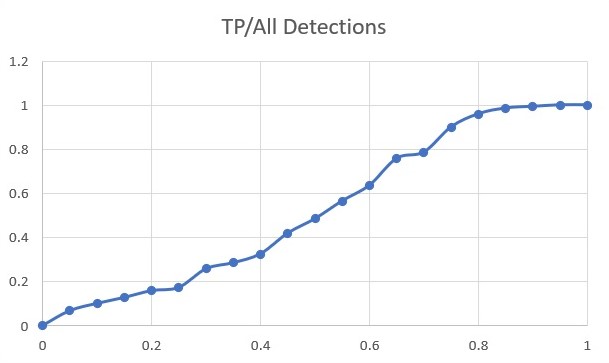}\\
(a)& (b)\\
\end{tabular}
\caption{(a) The accumulative number of "True Positives" and "False Positives" according to the detection score (b) The ratio of "True Positives" to all detections in each score range}
\label{fig:dets_plot}
\end{figure*}

The detection algorithm additionally reports a confidence score for each detection that shows how likely the detection is true. So, it is expected that detections with the highest score correspond to real targets. Evidently, as the score of detection decreases, it is more likely that the detection is a false positive. In Fig.~\ref{fig:dets_plot}(a) the accumulative number of true positives and false positives according to their score for sequence 10 of the MOT17~\cite{milan2016mot16} dataset is presented. In Fig.~\ref{fig:dets_plot}(b) the ratio of true positive to all detections in each score range is plotted. As it can be seen, most high score detections are true positives and the ratio of true positives to all detections is close to one. As the confidence score decreases, the number of true positives and the ratio decreases too. For scores below 0.1, this ratio comes near to zero. This suggests the tracking algorithm faces with high scored detections differently than low scored detections. In this paper, a high score and a low score threshold are defined in the tracking algorithm and matching is performed in a cascade manner. In the first step, the minimum similarity for matching is set near $0$ to increase the association between detections and tracks. In the second step, detections with lower confidence , which are more probable to be false positives, are used for association. Thus, the minimum similarity is set to a higher number to only highly similar detections match with the tracks that are not matched in the first step.

To compute similarity, tracking algorithms usually use one or a combination of motion, appearance, and interaction models. A motion model uses geometric features of bounding boxes such as size, position, and velocity of bounding boxes. An appearance model uses the color values of pixels inside the bounding box. This imposes a higher computational cost for using appearance cues. The appearance model is more effective for detecting occlusion and re-identifying occluded tracklets after reappearance. In the proposed algorithm pay attention to the relation of tracklets to detect occlusion and to help re-identify occluded tracklets effectively with low computational cost.

In the motion model, it is assumed that camera motion is negligible. But in practice, there is significant camera motion in some frames that mislead the tracking algorithm and causes some tracklets do not match or some of them match with the wrong detection. In our algorithm, we introduce a method for measuring the camera motion and removing its effect in data association that significantly improves the tracking performance in the presence of camera motion.

In summary, our contributions are:
\begin{itemize}
    \item a novel and fast online relational multiple object tracking algorithm
    \item a cascade association based on the detection score
	\item a novel interaction model based on geometric features for occlusion detection and target re-identification
    \item an algorithm to estimate camera motion to decrease the non-linearity motion of targets
    \item a new similarity metric which considers detection's size
    
\end{itemize}

\section{Related Work}
\label{sec:relatedwork}
In tracking-by-detection~\cite{bewley2016simple,wojke2017simple,yu2016poi,papakis2020gcnnmatch} paradigm, first a deep learning-based detection algorithm ~\cite{felzenszwalb2008discriminatively,ren2015faster,yang2016exploit,he2017mask,cai2018cascade,sun2021sparse,redmon2018yolov3} generates detections which are input of  the tracking algorithm. In the joint detection and tracking paradigm~\cite{lu2020retinatrack,shan2020tracklets,bergmann2019tracking,feichtenhofer2017detect}, the detections are extracted as the first step in the tracking algorithm and the features computed in this step can be used in tracking.
With having detections, the next step is to compute the similarity between tracks and detections. For computing similarity, geometric features, appearance features, and interaction and relational features can be used. Some simple methods such as SORT~\cite{bewley2016simple} only use position and motion features. Some other methods such as DeepSort~\cite{wojke2017simple} use appearance alongside geometric features and use deep neural networks for extracting appearance features. Although the results of the DeepSort algorithm become less attractive after a few years, recently in~\cite{du2022strongsort} a new deep neural network is used for extracting appearance features and could achieve near the state of the art results. The most drawback of using appearance features is the high computational cost that decreases the speed of the algorithm and makes it hard to use the algorithm for online applications.

In recent years, more attention is paid to the interaction features alongside other features. The graph convolutional neural network is used by a category of algorithms to model the interaction of tracks in spatial-temporal domains. In~\cite{papakis2020gcnnmatch} tracklets and detections are considered as nodes of a graph and if the similarity between a tracklet and a detection is higher than a threshold, an edge is created to connect the detection and tracklet. These graphs which are generated for each frame are the input of GCNN. The geometry and appearance features of nodes interact to generate more discriminative features. In ~\cite{wangjoint} the GNN is used in a joint paradigm to detect objects and associate them simultaneously. Also in~\cite{weng2020gnn3dmot} a GNN is used for feature extraction to fuse the 2D and 3D features of objects.

\begin{figure*}[!htb]
\centering
\includegraphics[width=16cm, height=7.3cm]{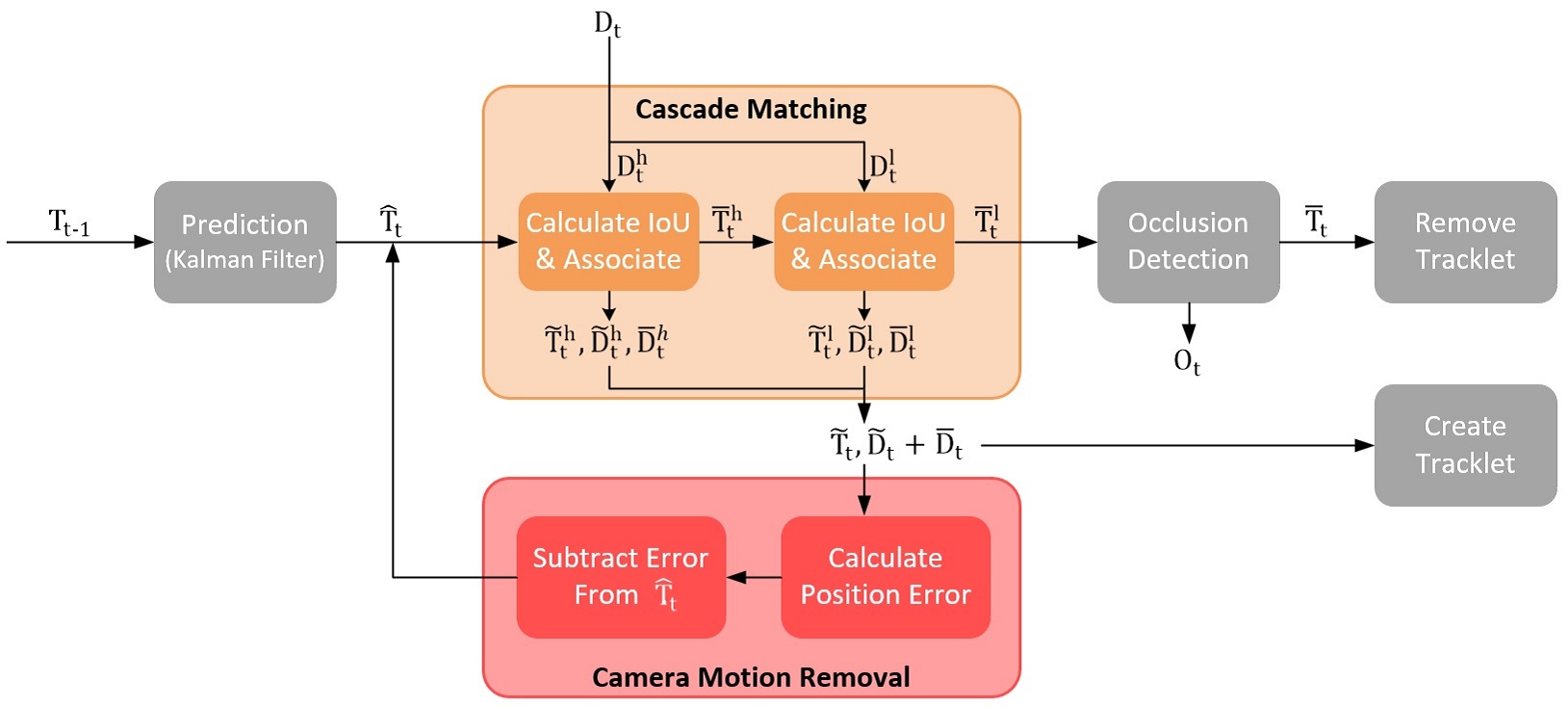}\\
\caption{The block diagram of the proposed algorithm}
\label{fig:block}
\end{figure*}

The transformers were first used in~\cite{vaswani2017attention} for natural language processing. But after that, some researchers used the transformer to model the interaction of objects and used in the tracking by attention paradigm. In~\cite{meinhardt2021trackformer} an end-to-end algorithm is presented which detects objects and associates them to tracklets by formulating MOT as a set prediction problem and association is performed through the concept of track queries. In~\cite{zeng2021motr} another end-to-end algorithm using transformers is presented. The~\cite{xu2021transcenter} uses the concept of tracking the center of objects which was introduced in ~\cite{zhou2020tracking} and a transformer-based architecture is used for tracking the center of targets. In~\cite{sun2020transtrack} a joint method for detection and association is presented using the transformer to share the knowledge of detection in re-identification. Also, in~\cite{chu2021transmot} the graph used in the transformer to model the interaction of objects. In this algorithm, a cascade association framework is used that first associate easy cases with only using Intersection over Union (IoU) and then use the transformer to associate hard cases. The cascade association in this algorithm decreases the computation cost.

recently, an algorithm is presented in ~\cite{zhang2021bytetrack} which only uses geometric features. In this work, low confidence detections are used to improve the results. In this paper, the detections of many recent papers ~\cite{zhou2020tracking,sun2020transtrack,wang2020towards,liang2022rethinking,pang2021quasi} are used as input to the presented tracking algorithm and show that their algorithm produces state-of-the-art results. 

In this paper, a novel method for modeling the interaction of tracklets is presented by improving the basic concept of the covered percent which is introduced in~\cite{nasseri2021simple}. This parameter is used for occlusion detection and it helps to keep the lost tracklets more efficiently for re-identification.  

\section{Approach}
In the proposed algorithm, we assume that two inputs are available at each frame. The first is a set of $K$ detections extracted by a detection algorithm at frame $t$ which are shown as $\mathbf{D_t}=\{D^1_t, D^2_t, ..., D^{K_t}_t\}$. The second input is a set of \textit{tracklets} created till frame $t-1$, where each tracklet is itself a set of detections assigned with an ID (i.e, target). The set of tracklets is represented by $\mathbf{T_{t-1}}=\{T^1_{t-1}, T^2_{t-1}, ...,$ $ T^{P_{t-1}}_{t-1}\}$. The diagram of the proposed algorithm is depicted in Fig.~\ref{fig:block} and the pseudo-code of it is written in Algorithm ~\ref{alg:algorithm}. In the following the details of each component is explained.

\subsection{Prediction}

As a first step, the position of tracklets is predicted at the current frame using a motion model. Here, the kalman filter ~\cite{kalman1960new} with a constant velocity motion model is used for prediction. The states of this model are:
\begin{equation}
x=[u, v, a, h, \dot{u}, \dot{v}, \dot{a}, \dot{h}]
\end{equation}
in which $u$ and $v$ are the center, $a$ is the aspect ratio and $h$ is the height of a tracklet box and the next four parameters are their corresponding changing rate. The predicted location of tracklets in the current frame is shown as $\mathbf{\hat{T}_t}$.

The detections and the prediction of tracklets in the current frame are the inputs of the association part of the proposed algorithm. after association, most of detections are assigned to their corresponding tracklets. These detections are now represented by $\mathbf{\tilde{D}_t}$ and the modified tracklets are represented by $\mathbf{\tilde{T}_t}$. Consequently, there are a few unmatched detections and tracklets which are represented by $\mathbf{\bar{D}_t}$ and $\mathbf{\bar{T}_t}$. 

\begin{figure*}[t]
\centering
\begin{tabular}{c c}
\includegraphics[width=6.3cm, height=4.5cm]{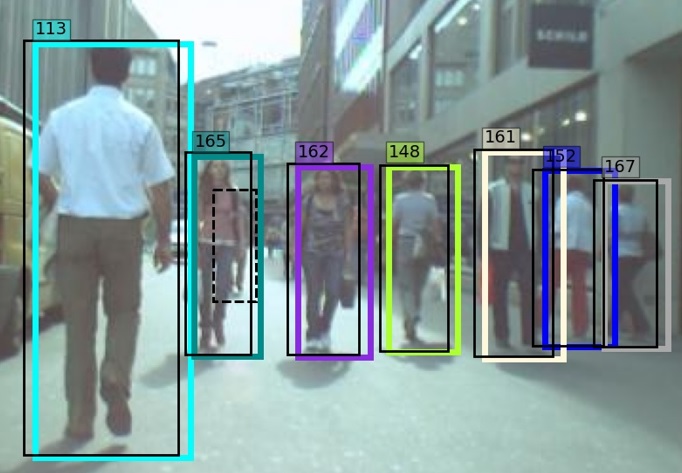}&
\includegraphics[width=6.3cm, height=4.5cm]{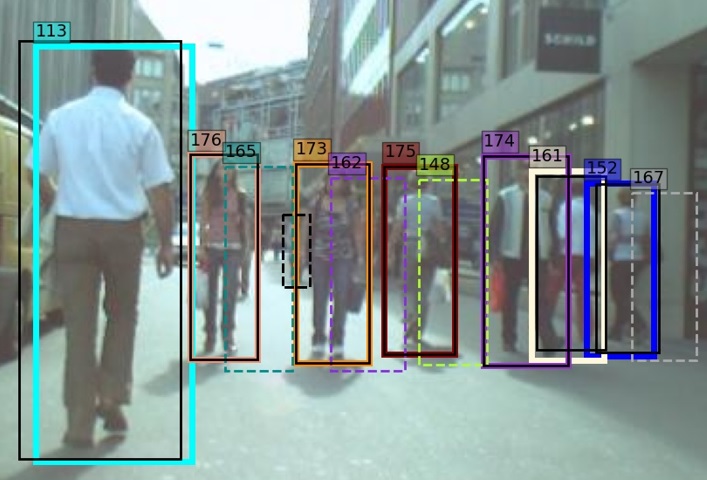}\\
(a)& (b)\\
\end{tabular}
\begin{tabular}{c c}
\includegraphics[width=6.3cm, height=4.5cm]{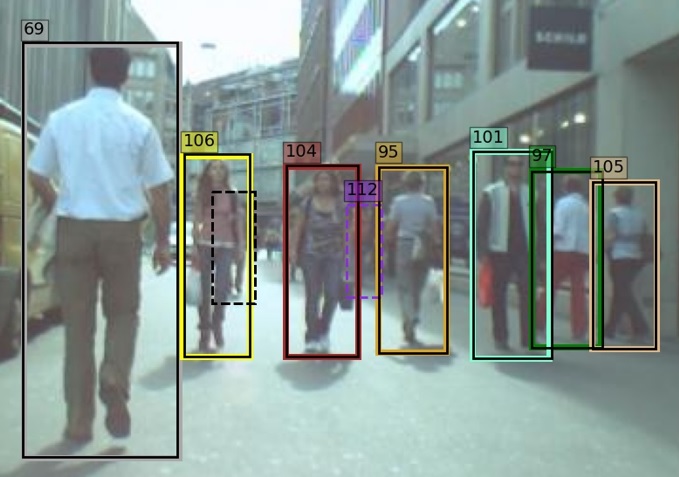}&
\includegraphics[width=6.3cm, height=4.5cm]{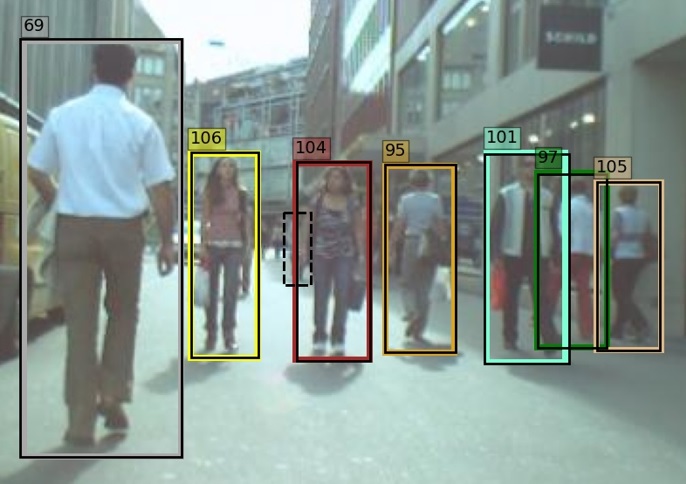}\\
(c)& (d)\\
\end{tabular}
\caption{The effect of camera motion removal in two consecutive frames with significant camera motion. The trackers bounding
boxes are shown with different color and the trackers ID is depicted in top of bounding box. In top figures the camera motion
removal is not applied. In bottom figures the camera motion removal is applied. (a) The trackers in frame t without camera
motion removal (b) The trackers in frame t+1 without camera motion removal (c) The trackers in frame t with camera motion
removal (d) The trackers in frame t+1 with camera motion removal}
\label{fig:cam_mot_rem}
\end{figure*}

\subsection{Cascade Matching}
For assigning detections to tracklets, an association metric based on the geometric features of bounding boxes are used to reduce the calculation cost. The proposed metric is the so-called \textit{normalized Intersection over Union}(nIoU). To compute the $nIoU$, we first need to compute the normalized difference between position and size of boxes:

\begin{equation}
\begin{aligned}
  d_{u} = \displaystyle \frac{abs(u_{D^i_t}-u_{\hat{T}^j_t})}{w_{D^i_t}}\\
  d_{v} = \displaystyle \frac{abs(v_{D^i_t}-v_{\hat{T}^j_t})}{h_{D^i_t}}\\ 
  d_{w} = \displaystyle \frac{abs(w_{D^i_t}-w_{\hat{T}^j_t})}{w_{D^i_t}}\\
  d_{h} = \displaystyle \frac{abs(h_{D^i_t}-h_{\hat{T}^j_t})}{h_{D^i_t}}
\end{aligned}
\end{equation}

Now, the average of these four terms is subtracted from IoU to get normalized IoU:
\begin{equation}
nIoU = \frac{I(A_{D^i_t}, A_{T^j_t})}{A_{D^i_t}+A_{T^j_t}-I(A_{D^i_t}, A_{T^j_t})}-\frac{(d_{u} + d_{v} + d_{w} + d_{h})}{4}
\end{equation}
in which $A_{D^i_t}$ and $A_{T^j_t}$ are the area of $i^th$ detection box and predicted detection box of $j^th$ tracklet in current frame.  Also, the $I$ function calculates the intersection of these two bounding boxes.  
From detection algorithm, a confidence score for each detection is also reported that shows the probability of existence of a track. In Fig.~\ref{fig:dets_plot} detections along with their corresponding score is visualized. Evidently, as the detection score decreases, the chance of track's existence decreases. Therefore, our tracking algorithm considers this cue in matching detections with tracklets in two steps. Here, the tracking algorithm tries to match most high score detections and match low score detections with more caution. To this end, two thresholds on detection score are defined as $L_{h}$ and $L_{l}$. First, detections with a score higher than $L_{h}$ are matched with tracklets. In this step, the minimum accepted nIoU is set at a low value to match high score detections whether there is a minimum similarity. The detections with a score between these two thresholds enter the second matching step and are matched with the tracklets which are not matched in the first step. Because the chance of existence of false-positive detections is higher in this step, matching is done with more caution by increasing the minimum accepted similarity. At the end, detections whose score is below than $L_{l}$ are discarded as with high probability they are false positives. The tracklets which are not matched in the second step are reported as unmatched tracklets. The other variables of the two steps are aggregated and reported as the corresponding variable. The pseudo-code of this block is shown in Algorithm ~\ref{alg:functions}.

\begin{algorithm}[!ht]
\caption{Proposed Algorithm}
\DontPrintSemicolon
\LinesNumbered
\SetKwProg{Fn}{Function}{}
\KwData{\textbf{Data:} $D_{t}$, $T_{t-1}$}\;
\KwResult{$T_{t}$}
$\hat{T}_t = KF\_Prediction(T_{t-1})$\; \label{alg:algorithm}
$\tilde{T}_t, \tilde{T}_t, \bar{T}_t, \bar{D}_t = Cascade\_Matching(D_t, \hat{T}_t)$\;
\textbf{Camera Motion Removal:}\;
$u_{avg} = avg(abs(u_{\tilde{T}_t}-u_{\tilde{T}_t}))$\;
$v_{avg} = avg(abs(v_{\tilde{T}_t}-v_{\tilde{T}_t}))$\;
$u_{\hat{T}_t} = u_{\hat{T}_t} - u_{avg}$\;
$v_{\hat{T}_t} = v_{\hat{T}_t} - v_{avg}$\;
$\tilde{T}_t, \tilde{T}_t, \bar{T}_t, \bar{D}_t = Cascade\_Matching(D_t, \hat{T}_t)$\;
$\tilde{T}_t=KF\_Correction(\tilde{T}_t, \tilde{D}_t)$\;
$O_t, \hat{T}_t = Occlusion\_Detection(\hat{T}_t)$\;
$\hat{T}_t = Remove\_Tracklets(\hat{T}_t)$\;
$T^n_t = Create\_Tracklets(\hat{D}_t)$\;
$T_{t} \leftarrow \tilde{T}_t + O_t + T^n_t + \bar{T}_t$\;
\end{algorithm}

\subsection{Camera Motion Removal}
In constant velocity motion model, the assumption is that targets moves with the constant velocity and the camera motion is negligible. But this is a strong assumption which is not always valid. Sometimes the camera motion is so significant that causes challenges in matching. When the camera moves, detections move in the opposite direction and the amount of their movement is proportional to the camera movement. In such cases, the camera motion influences the matching of detections with tracklets and increases the risk of ID switch or mismatching. This effect is even sever for the tracks who are far from the camera and their bounding boxes are small. 

In the proposed algorithm, we introduce a novel method for measuring camera motion and removing it's effect on association. When the camera motion is high, all tracks have displacement relative to their predicted position. So, after matching, the difference between the position of matched pair of detection-tracklet is calculated in each direction. These differences are averaged and reported as the amount of camera motion. This value is subtracted from the position of all tracklet boxes and the matching step is repeated with the modified predicted position of tracklets. In Fig.~\ref{fig:cam_mot_rem} the effect of the proposed method for two consecutive frames with significant camera motion is shown. In the top figures, the camera motion removal is not applied. As can be seen, some tracklets are not matched or matched incorrectly. At the bottom, the camera motion removal is applied. As it can be seen, the identity of tracklets is preserved correctly.  

\begin{algorithm}[!ht]
\caption{Algorithm Functions}
\DontPrintSemicolon
\LinesNumbered
\SetKwProg{Fn}{Function}{}
\KwData{}
\KwResult{}
\Fn{Cascade\_Matching($D_t, \hat{T}_t$)}{
\For{$d \in D_t$}{\label{alg:functions}
\If{$s_d > L_h$}{
$D^h_t \leftarrow d$\;
} 
\uElseIf{$s_d > L_1$}{
$D^l_t \leftarrow d$\;
} 
} 
$P^h = nIoU(\hat{T}_t, D^h_t)$\;
$\tilde{T}^h_t, \tilde{D}^h_t, \bar{T}^h_t, \bar{D}^h_t = Hungarian(P^h)$\;
$P^l = nIoU(\bar{T}^h_t, D^l_t)$\;
$\tilde{T}^l_t, \tilde{D}^l_t, \bar{T}^l_t, \bar{D}^l_t = Hungarian(P^l)$\;
$\tilde{D}_t \leftarrow \tilde{D}^h_t + \tilde{D}^l_t$\;
$\tilde{T}_t \leftarrow \tilde{T}^h_t + \tilde{T}^l_t$\;
$\bar{D}_t \leftarrow \bar{D}^h_t + \bar{D}^l_t$\;
$\bar{T}_t \leftarrow \bar{T}^l_t $\;
} 
\;
\Fn{Occlusion\_Detection($\bar{T}_t, \hat{T}_t$)}{
$O_t \leftarrow \emptyset$\;
$A_{avg} = Average(A_{\hat{T}_t})$\;
$R = CR(\hat{T}_t)$\;
\For{$ u \in \bar{T}_t$}{
$A^n_u = \frac{A_{u}}{A_{avg}}$\;
$C_u = \frac{Age_{u}}{t_{so_{u}}}*A^n_u$\;
\If{$R_u > L_{cr} \textbf{and} C_u > L_c$}{
$O_t \leftarrow O_t + u$\;
$\bar{T}_t \leftarrow \bar{T}_t - u$\;
} 
} 
} 
\;
\Fn{Remove\_Tracklets($\bar{T}_t, \hat{T}_t$)}{
$R = CR(\hat{T}_t)$\;
\For{$ u \in \bar{T}_t$}{
\If{$R_u > L_{cr}$}{
$t_{snc_u} \leftarrow t_{snc_u} + 1$\;
} 
\If{$t_{snc_u} > 2$}{
$\bar{T}_t \leftarrow \bar{T}_t - u$\;
} 
} 
} 
\;
\Fn{Create\_Tracklets($\bar{D}_t$)}{
$T^n_t \leftarrow \emptyset$;\;
\For{$ d \in \bar{D}_t$}{
\If{$s_d > L_{n}$}{
$T^N_t \leftarrow T^n_t + d$\;
} 
} 
} 
\end{algorithm}

\begin{table*}[th!]
\centering
\begin{tabular}{l c c c c c c c c c}
\hline
\textbf{Tracker} & \textbf{MOTA} $\uparrow$ & \textbf{IDF1} $\uparrow$ & \textbf{HOTA} $\uparrow$ & \textbf{MT} $\uparrow$ & \textbf{ML} $\downarrow$ & \textbf{FP} $\downarrow$ & \textbf{FN} $\downarrow$ & \textbf{IDS} $\downarrow$ & \textbf{FM} $\downarrow$\\
\hline
FairMOT~\cite{zhang2021fairmot} & 73.7 & 72.3 & 59.3 & 1017 & 408 & 27507 & 117477& 3303 & 8073\\
GCNNMatch~\cite{papakis2020gcnnmatch} & 57.3 & 56.3 & 45.4 & 575 & 787 & \textbf{14100} & 225042 & \textbf{1911} & 2837\\
GSDT~\cite{wangjoint} & 73.2 & 66.5 & 55.2 & 981 & 411 & 26397 & 120666 & 3891 & 8604\\
Trackformer~\cite{meinhardt2021trackformer} & 74.1 & 68.0 & 57.3 & 1113 & 246 & 34602 & 108777 & 2829 & 4221\\
TransTrack~\cite{sun2020transtrack} & 75.2 & 63.5 & 54.1 & 1302 & \textbf{240} & 50157 & 86442 & 3603 & 4872\\
TransCenter~\cite{xu2021transcenter} & 73.2 & 62.2 & 54.5 & 960 & 435 & 23112 & 123738 & 4614 & 9519\\
TransMOT~\cite{chu2021transmot} & 76.7 & 75.1 & 61.7 & 1200 & 387 & 36231 & 93150 & 2346 & 7719 \\
ReMOT~\cite{yang2021remot} & 77.0 & 72.0 & 59.7 & 1218 & 324 & 33204 & 93612 & 2853 & 5304\\
\hline
ByteTrack~\cite{zhang2021bytetrack} & 80.3 & 77.3 & 63.1 & 1254 & 342 & 25491 & 83721 & 2196 & 2277 \\
OCSORT~\cite{cao2022observation} & 78.0 & 77.5 & 63.2 & 966 & 492 & 15129 & 107055 & 1950 & \textbf{2040}\\
Ours & \textbf{80.4} & \textbf{77.7} & \textbf{63.3} & \textbf{1311} & 255 & 28887 & \textbf{79329} & 2325 & 4689\\ 
\hline
\end{tabular}
\caption{Results on MOT17 test set with the private detections. ByteTrack, OCSORT and Ours use the same detections. The best results are shown in bold.}
\label{tab:res_mot17}
\end{table*}

\begin{table*}[th!]
\centering
\begin{tabular}{l c c c c c c c c c}
\hline
\textbf{Tracker} & \textbf{MOTA} $\uparrow$ & \textbf{IDF1} $\uparrow$ & \textbf{HOTA} $\uparrow$ & \textbf{MT} $\uparrow$ & \textbf{ML} $\downarrow$ & \textbf{FP} $\downarrow$ & \textbf{FN} $\downarrow$ & \textbf{IDS} $\downarrow$ & \textbf{FM} $\downarrow$\\
\hline
FairMOT~\cite{zhang2021fairmot} & 61.8 & 67.3 & 54.6 & 855 & 94 & 103440 & 88901& 5243 & 7874\\
GCNNMatch~\cite{papakis2020gcnnmatch} & 54.5 & 49.0 & 40.2 & 407 & 317 & \textbf{9522} & 223611 & 2038 & 2456\\
GSDT~\cite{wangjoint} & 67.1 & 67.5 & 53.6 & 660 & 164 & 31507 & 135395 & 3230 & 9878\\
Trackformer~\cite{meinhardt2021trackformer} & 68.6 & 54.7 & 65.7 & 666 & 181 & 20348 & 140373 & 1532 & 2474\\
TransTrack~\cite{sun2020transtrack} & 65.0 & 59.5 & 48.9 & 622 & 167 & 27191 & 150197 & 3608 & 11352\\
TransCenter~\cite{xu2021transcenter} & 61.0 & 49.8 & 43.5 & 601 & 192 & 49189 & 147890 & 4493 & 8950\\
TransMOT~\cite{chu2021transmot} & 77.5 & 75.2 & 61.9 & \textbf{878} & \textbf{113} & 34201 & \textbf{80788} & 1615 & 2421 \\
ReMOT~\cite{yang2021remot} & 77.4 & 73.1 & 61.2 & 846 & 123 & 28351 & 86659 & 1789 & 2121\\
\hline
ByteTrack~\cite{zhang2021bytetrack} & \textbf{77.8} & 75.2 & 61.3 & 859 & 118 & 26249 & 87594 & 1223 & 1460 \\
OCSORT~\cite{cao2022observation} & 75.7 & 76.3 & \textbf{62.4} & 813 & 160 & 19067 & 105894 & \textbf{942} & \textbf{1086}\\
Ours & 76.8 & \textbf{76.4} & 61.4 & 856 & 133 & 27106 & 91740 & 1446 & 3053\\ 
\hline
\end{tabular}
\caption{Results on MOT20 test set with the private detections. ByteTrack, OCSORT and Ours use the same detections. The best results are shown in bold.}
\label{tab:res_mot20}
\end{table*}

\subsection{Occlusion Handling}
One of the challenges in multiple object tracking algorithms is where a tracklet is covered partially or completely by other objects or tracklets and is not observed by the camera. In~\cite{zhang2021bytetrack} the low score detections are used to find partially seen tracklets between them. As explained in the previous sections, the probability of false positives between these detections is high. So, in our approach for detecting occlusion, a method based on the relation of tracklets is developed. As the tracklet boxes are known, occlusion can be detected by considering their position relative to each other. To quantify the relation of them, the tracklet covered ratio is defined based on the similar concept in ~\cite{nasseri2021simple} as:
\begin{equation}
CR_{i} = \max_{\{T^j_t \in {T}_t\}}\frac{I(A_{T^i_t},A_{T^j_t})}{A_{T^i_t}}
\end{equation}
This parameter shows the maximum ratio of a tracklet box that is covered by any of the other tracklets. The difference between this parameter and IoU is that the intersection of two bounding boxes is divided into the area of one bounding box, instead of the union of two bounding boxes. The high amount of this parameter can point to the occurrence of occlusion. To improve the precision of occlusion detection, another parameter is used alongside this as the tracklet's confidence:
\begin{equation}
C_i = \frac{Age_{i}}{t_{so_i}} * A^n_i 
\label{eq:conf}
\end{equation}
in which $Age_{i}$ is the total number of frames that tracklet is existed, $t_{so_i}$ or time since observed is the number of successive frames the tracklet is not matched with any detection and $A^n_i$ is the normalized area of tracklet box which is calculated by division of area of tracklet box to the average area of all tracklet boxes. This parameter measures the confidence of a reported tracklet by tracking algorithm. When a tracklet is tracked for more frames and is nearer to the camera and seen bigger, the probability of its existence is higher. On the other hand, when a tracklet is not visible for more frames, this confidence is decreased. 

For the tracklets that do not match in the matching step, these two above parameters are calculated. If $CR_i$ is higher than $L_{cr}$ and $C_i$ is higher than $L_c$, that tracklet is reported as occluded in $\mathbf{O_t}$.

\begin{table*}[th!]
\centering
\begin{tabular}{l c c c c c c c c c}
\hline
\textbf{Sequence} & \textbf{Method} & \textbf{MOTA} $\uparrow$ & \textbf{IDF1} $\uparrow$ & \textbf{MT} $\uparrow$ & \textbf{ML} $\downarrow$ & \textbf{FP} $\downarrow$ & \textbf{FN} $\downarrow$ & \textbf{IDS} $\downarrow$ & \textbf{FM} $\downarrow$\\
\hline
Sequence 5 & Without CMR & 81.0\% & 75.9\% & 95 & 6 & 444 & 691 & 177 & 134\\
Sequence 5 & With CMR & 82.7\% & 78.3\% & 98 & 6 & 460 & 666 & 73 & 122\\
\hline
Sequence 10 & Without CMR & 77.5\% & 59.5\% & 57 & 0 & 673& 1976 & 237 & 298\\
Sequence 10 & With CMR & 82.6\% & 75.5\% & 46 & 0 & 663 & 1486 & 89 & 273\\
\hline
\end{tabular}
\caption{The effect of camera motion algorithm on two sequences of MOT17 dataset with high camera motion}
\label{tab:cmr_eff}
\end{table*}

\begin{table*}[th!]
\centering
\begin{tabular}{l c c c c c c c c c}
\hline
\textbf{Method} & \textbf{MOTA} $\uparrow$ & \textbf{IDF1} $\uparrow$ & \textbf{MT} $\uparrow$ & \textbf{ML} $\downarrow$ & \textbf{FP} $\downarrow$ & \textbf{FN} $\downarrow$ & \textbf{IDS} $\downarrow$ & \textbf{FM} $\downarrow$\\
\hline
Ground truth matching & 90.5\% & 95.0\% & 391 & 18 & 0 & 10697 & 0 & 1856\\
Proposed algorithm & 90.7\% & 85.0\% & 442 & 21 & 2947 & 7031 & 488 & 857\\
\hline
\end{tabular}
\caption{The effect of occlusion detection algorithm with comparison of ground-truth matching results with the proposed algorithm results on MOT17 dataset}
\label{tab:od_eff}
\end{table*}

\subsection{Removing Tracklets}
The unmatched tracklets which are not detected as occluded are possible candidates for removal. In some works~\cite{bewley2016simple,wojke2017simple,du2022strongsort,zhang2021bytetrack}, they are kept for a certain number of frames (L frames). If L is a low number such as 3, the occluded tracks after reappearing will get a new identity giving rise to an ID Switch. If a high number such as 30 is chosen for L, the number of kept tracklets increases that causes a high computation cost. The lost tracklets are kept for occlusion cases to the identity of occluded tracks is preserved after reappearing. The target covered ratio ($CR$) which is used for occlusion detection in the proposed algorithm, is used for keeping lost tracklets, too. While the covered ratio of a tracklet is higher than $L_{cr}$ it is considered as covered. But when the covered ratio of that tracklet comes below this threshold for a few consecutive frames, i.e. 3, it is removed.  

\subsection{Creating Tracklets}
Detections that are not matched with any tracklets may point to existence of new tracklets. If their score is higher than $L_n$ which is close to 1, the probability that they indicate a real tracks is very high. So, for each detection a tracklet is created initialized with that detection. Other unmatched detections are kept and passed for association in the next frame.

\section{Experiments}
To evaluate the performance of our proposed algorithm and compare its results with other tracking algorithms, the MOT17 ~\cite{milan2016mot16} and MOT20 ~\cite{dendorfer2020mot20} datasets are used with private detections. The detections are same as ~\cite{zhang2021bytetrack} and extracted using YOLOX ~\cite{ge2021yolox} algorithm.

\subsection{Metrics}
For comparison of result, the widely used CLEAR MOT metrics ~\cite{bernardin2008evaluating} which include Multiple Object Tracking Accuracy (MOTA), Multiple Object Tracking Precision (MOTP) are reported. In addition, the HOTA ~\cite{luiten2021hota} and IDF1 ~\cite{ristani2016performance} are reported. The MOTA metric consider FP, FN and IDS effect equally. As the number of FP and FN are much higher than IDS, the focus of this metric is more on detection performance. But in HOTA and IDF1 more attention is paid to association performance by focusing on preservation of track's identity. Also, other popular metrics are used including Mostly Tracked (MT), Mostly Lost (ML), the number of False Negatives (FN), False Positives (FP), ID-Switches (IDS), and track Fragmentation (FM).

\subsection{Results}
The results of running the proposed algorithm on MOT17 and MOT20 datasets are presented in ~\ref{tab:res_mot17} and ~\ref{tab:res_mot20} respectively. The result of other state-of-the-art algorithms, especially the ones using interaction models are also available in these tables for comparison. All of these algorithms use private detections. The private detections that we used are the same as two other algorithms ~\cite{zhang2021bytetrack,cao2022observation}. So, the results of our algorithm and these two algorithms are separated by a line at bottom of the tables. In MOT17, our algorithm achieves the best results on MOTA, IDF1, HOTA, and, MT among other algorithms. Also in MOT20, the results of our algorithm are comparable with the state-of-the-art algorithms.  

\subsection{Camera Motion Removal Effectiveness}
In the proposed algorithm, a method for removing camera motion is presented. MOT17 dataset is composed of 7 training and 7 test sequences. In most sequences, the camera motion is negligible. So, the camera motion removal has no significant effect. among MOT17 training sequences, sequences 5 and 9 have significant camera motion in several frames. The results of the proposed algorithm with and without camera motion removal block are presented in the table~\ref{tab:cmr_eff}. As it can be seen, the camera motion removal could notably improve MOTA, IDF1, and ID Switch metrics.

\subsection{Occlusion Detection Effectiveness}
The occlusion detection method in the proposed algorithm calculates the covered ratio and confidence of tracklets that are not matched with any detections and report them as tracklet of that frame by checking some conditions on these parameters. So, the tracklets that are marked as occluded, do not have any corresponding detection. To investigate the effectiveness of this block, the detections in training sequences are matched with ground truths. In this way, all the detections match correctly with the corresponding tracks, and the number of false positives and ID switches reaches zero and the best possible result is achieved. In this way no track will be reported when it has no corresponding detection. The results in this situation with the result of the proposed algorithm on the training set are presented in the table ~\ref{tab:od_eff}. As it can be seen, the proposed algorithm could pass the best achievable MOTA by reporting some unmatched tracklets as occluded. 

\section{Conclusion}
In the proposed algorithm a novel interaction model based on geometric features is presented. This model considers the relation of targets to detect occlusion and re-identify lost targets. In comparison to other algorithms with interaction models that are using GCNN or Transformer, the presented algorithm has a lower computational cost. In addition, we introduced a method for measuring camera motion and removing its effect on data association. This method could effectively improve the tracking result in presence of camera motion. Our approach achieves state-of-the-art results on the MOT17 and comparable results on the MOT20 dataset, while it is super fast that makes it ideal for real-time applications.

{\small
\bibliographystyle{ieee_fullname}
\bibliography{ForTracking}
}

\end{document}